# AN OPTIMIZED HYBRID APPROACH FOR PATH FINDING

Ahlam Ansari[1] and Mohd Amin Sayyed[2], Khatija Ratlamwala[2], Parvin Shaikh[2]

[1]Assistant Professor at M.H.Saboo Siddik College of Engineering, University of Mumbai, India
[2]Students of M.H.Saboo Siddik College of Engineering, University of Mumbai, India

## ABSTRACT

*Path finding algorithm addresses problem of finding shortest path from source to destination avoiding obstacles. There exist various search algorithms namely A\*, Dijkstra's and ant colony optimization. Unlike most path finding algorithms which require destination co-ordinates to compute path, the proposed algorithm comprises of a new method which finds path using backtracking without requiring destination co-ordinates. Moreover, in existing path finding algorithm, the number of iterations required to find path is large. Hence, to overcome this, an algorithm is proposed which reduces number of iterations required to traverse the path. The proposed algorithm is hybrid of backtracking and a new technique(modified 8-neighbor approach). The proposed algorithm can become essential part in location based, network, gaming applications. grid traversal, navigation, gaming applications, mobile robot and Artificial Intelligence.*

## 1. INTRODUCTION

The path finding analysis has importance for various working such as logistics, operation management, transportation, system analysis and design, project management, game programming, network and production line. The shortest path analysis developed capability about think cause and effect, learning and thinking like human in artificial intelligence [1]. The path finding algorithm aims to minimize the cost from source to destination. There are various algorithm proposed for path finding.  A\* algorithm is a heuristic based algorithm widely used in Artificial Intelligent it follows path having lowest known heuristic cost. Drawback of A\* is it requires a large amount of CPU resources that is memory. Ant colony algorithm is based on behaviour of ants. In ant colony algorithm nodes are traversed in random fashion initially and cost of each node in the path is updated. The path which is used maximum time by ant is considered to be optimal. The main drawback of this algorithm is number of iteration required is more. Proposed algorithm exploits good properties of ant colony algorithm and A\* algorithm to develop a new algorithm which is efficient than the original algorithms. The proposed algorithm traverses nodes in all eight directions of source node during each iteration and updates cost of each node which is the distance of that node from center. The algorithm uses traversal which is similar to flood fill algorithm. Flood fill algorithm is used to solve the robot maze problem[2]. The proposed algorithm is an optimized hybrid approach that can be used to solve robot maze problem. In addition to robot maze problem it can also find application in any problem that requires shortest path computation.

## 2. LITERATURE REVIEW

### 2.1. Genetic Algorithm

Genetic algorithm is used to find approximate optimal solution. It is inspired by evolutionary biology such as inheritance, mutation, crossover and selection [3]. Advantages of this algorithm are it solves problem with multiple solutions, it is very useful when input is very large. Disadvantages of GA are certain optimization problems cannot be solved due to poorly known fitness function, it cannot assure constant optimization response times, in GA the entire population is improving, but this could not be true for an individual within this population [4].

### 2.2. Heuristic Function

Heuristic function maps problem state descriptor to a number which represents degree of desirability. Heuristic function has different errors in different states. It plays vital role in optimization problem [5].

### 2.3. Depth-First-Search

DFS uses Last-In-First out stack and are recursive in algorithm. It is simple to implement. But major problem with DFS is it requires large computing power, for small increase in map size, runtime increases exponentially [5].

### 2.4. Breadth-First-Search

BFS uses First-In-First-Out queue. It is used when space is not a problem and few solutions may exist and at least one has shortest path. It works poorly when all solutions have long path length or there is some heuristic function exists. It has large space complexity [5].

### 2.5. A* algorithm

A* combines feature of uniform-cost search and heuristic search. It is BFS in which cost associated with each node is calculated using admissible heuristic [6]. For graph traversal, it follows path with lowest known heuristic cost. The time complexity of this algorithm depends on heuristic used. Since it is BFS drawback of A* is large memory requirement because entire open-list is to be saved [5].

### 2.6. Ant colony optimization

ACO is meta-heuristic algorithm based on the behavior of real ant [7]. While traversing to destination each ant deposits chemical substance called pheromone. Each ant's traverses in random fashion but when it encounters pheromone trail it has to decide whether to follow it or not. If ant follows same path than amount of pheromone deposition increases. Thus the path followed by ant most has maximum pheromone deposition. An ant using shortest path to destination will reach source fast, as shortest path will have twice pheromone deposition than others [7]. If there is more than one optimal path, then ACO cannot decide optimal path [9].

### 2.7. Flood Fill Algorithm

Robot maze problems are an important field of robotics and it is based on decision making algorithm [10]. It requires complete analysis of workspace or maze and proper planning [11]. Flood fill algorithm and modified flood fill are used widely for robot maze problem [2]. Flood fill algorithm assigns the value to each node which is represents the distance of that node from centre [9]. The flood fill algorithm floods the the maze when mouse reaches new cell or node. Thus it requires high cost updates [9]. These flooding are avoided in modified flood fill [2].

# 3. PROPOSED SYSTEM

Given a map, a source node and a destination node, a least cost path from source to destination is to be computed. The proposed system starts with source node and search in all eight directions of source. This process is continued until destination node is reached. In the process  cost of each node which is equivalent to number of iterations required to reach that node is stored. Now to find shortest-path backtracking is performed from destination node to source node by including nodes with minimum cost amongst other in the path.

## 3.1. Algorithm

STEP 1: Initialize source point
STEP 2: Starting from source check upper row in the range ((iteration*2) +1) and assign iteration number to the node if node is ".".
2.1: If the current node is "@" (blocked or hurdle), then check left and right neighbours of "@", till "." (Pass through) is not found or till range is exhausted.
2.2: If pass through is found mark it as new source a repeat from Step 2.
STEP 3: Starting from source check upper row in the range ((iteration*2) +1) and assign iteration number to   the node if node is ".".
3.1: If the current node is "@" (blocked or hurdle), then check left and right neighbours of "@", till "." (Pass through) is not found or till range is exhausted.
3.2: If pass through is found mark it as new source repeat from Step 3.
STEP 4: For down repeat Step 2.
STEP 5: For left repeat Step 3.
STEP 6: Repeat from Step 2 to Step 5 till we find boundary on all four-sides or we find destination.
STEP 7: If destination found check all neighbours of that node and select smallest weight node and store in path array.
7.1: If node has more than one candidate having same minimum cost, make separate array for every path.
STEP 8: Repeat Step 7 till source is found.
STEP 9: Display path.
STEP 10: End.

## 3.2. Working

The working of algorithm is illustrated using example. The example can also be considered as robot maze in which source and destination is defined and shortest path is to be computed avoiding obstacles. Consider Figure 1. as input to which algorithm is applied. The steps involved in solving this problem is describe in detail.

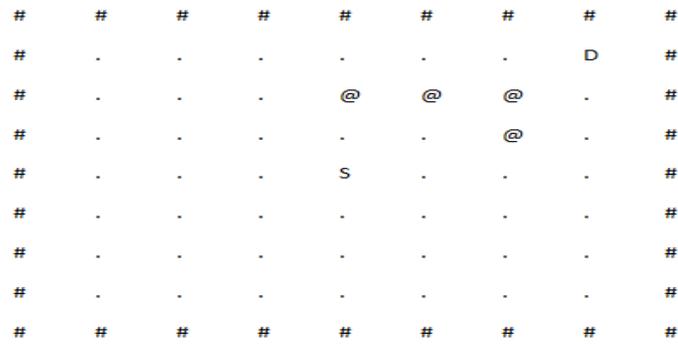

Figure 1. Input

In the above figure notations used are as follows:

\# : Boundary of the path.
@ : Block or hurdle
S: Source
D: Destination

The goal is to find shortest path from source to destination avoiding hurdles in the path. In order to do this the proposed algorithm follows following steps starting from source node:

- First iteration:

Traversing in upward direction of source node

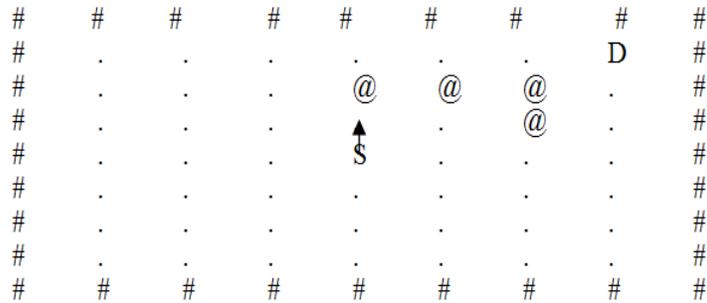

Figure 2. Traversing upwards

Similarly traversal in all 8-neighbouring point of source is done in first iteration as shown in Figure 3.

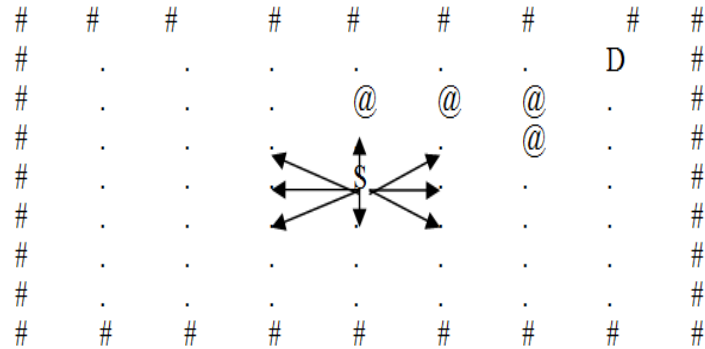

Figure 3. Traversal in fist iteration

As cost of each node is equal to number of iteration required to reach there,the cost of all these 8-neighbour will be updated as 1 as shown in figure 4.

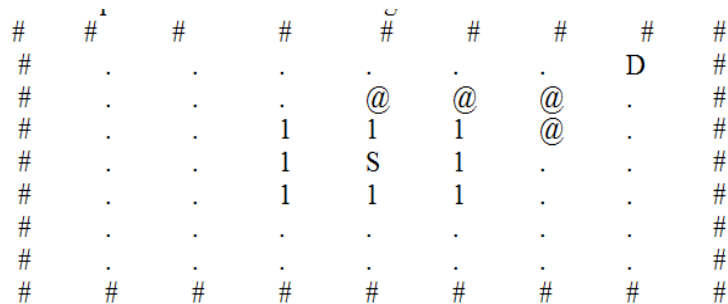

Figure 4. Cost updated after first iteration

• Second iteration:

In second iteration, the next level nearest neighbour of the source will be traversed as shown in figure 5.

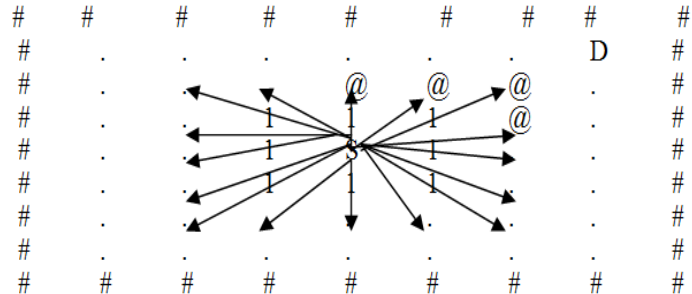

Figure 5. Traversal in second iteration.

The cost of the entire pass through "." will be updated as 2. In order to avoid obstacles from path the cost of blocked node is updated as infinity. Now from the blocked or "@" neighbours are checked till "." (Pass through) is found. These pass through node will marked as new source. As shown in figure 6.

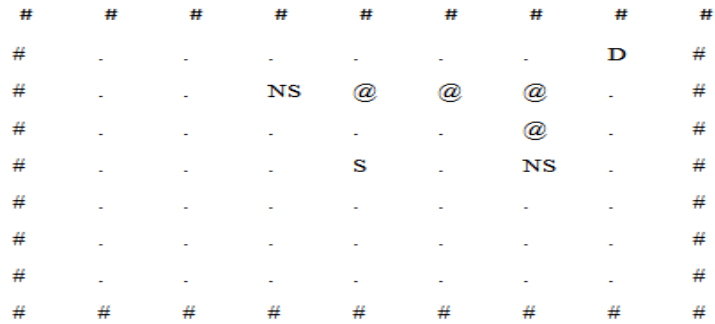

Figure 6. Marking of new source

In figure 6, NS represents new source. From these sources the traversal of node is in same fashion as from the original source (i.e. 8 nearest neighbour). But the traversal from these new sources will not update the cost of already updated node (i.e. nodes that are updated by original source). Moreover, the traversal from these sources as well as from original source is done simultaneously (not shown in the figure).

- Cost matrix after all iterations:

The cost matrix after all iterations (3 in this case) is shown below in figure 7.

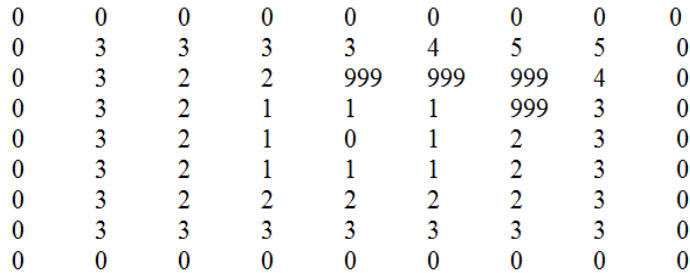

Figure 7. Cost Matrix

- Reverse traversal:
- 

In order to find shortest path reverse traversal is done from destination. All the neighbouring node of destination is checked and node with minimum cast is selected. This step is repeated until source is found. Figure 8 shows the nodes included in the shortest path.

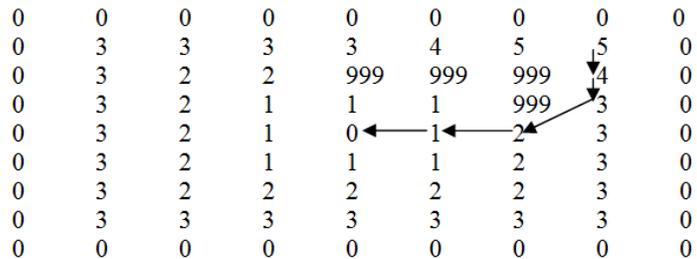

Figure 8. Finding of Shortest Path by Reverse traversal

### 3.3. Flowchart

Flowchart of map exploring i.e. path traversal and backtracking from destination to source to find shortest path is shown below.

Figure 9 shows flowchart of map exploring i.e. how path is traversed. In proposed algorithm traversal is done in all eight directions of source node in one iteration and cost of each node is updated. If boundary is found in all four directions and destination is not found, then it indicates that destination is not reachable. If an obstacle is found while traversing, then its cost is updated to a higher value (infinity).

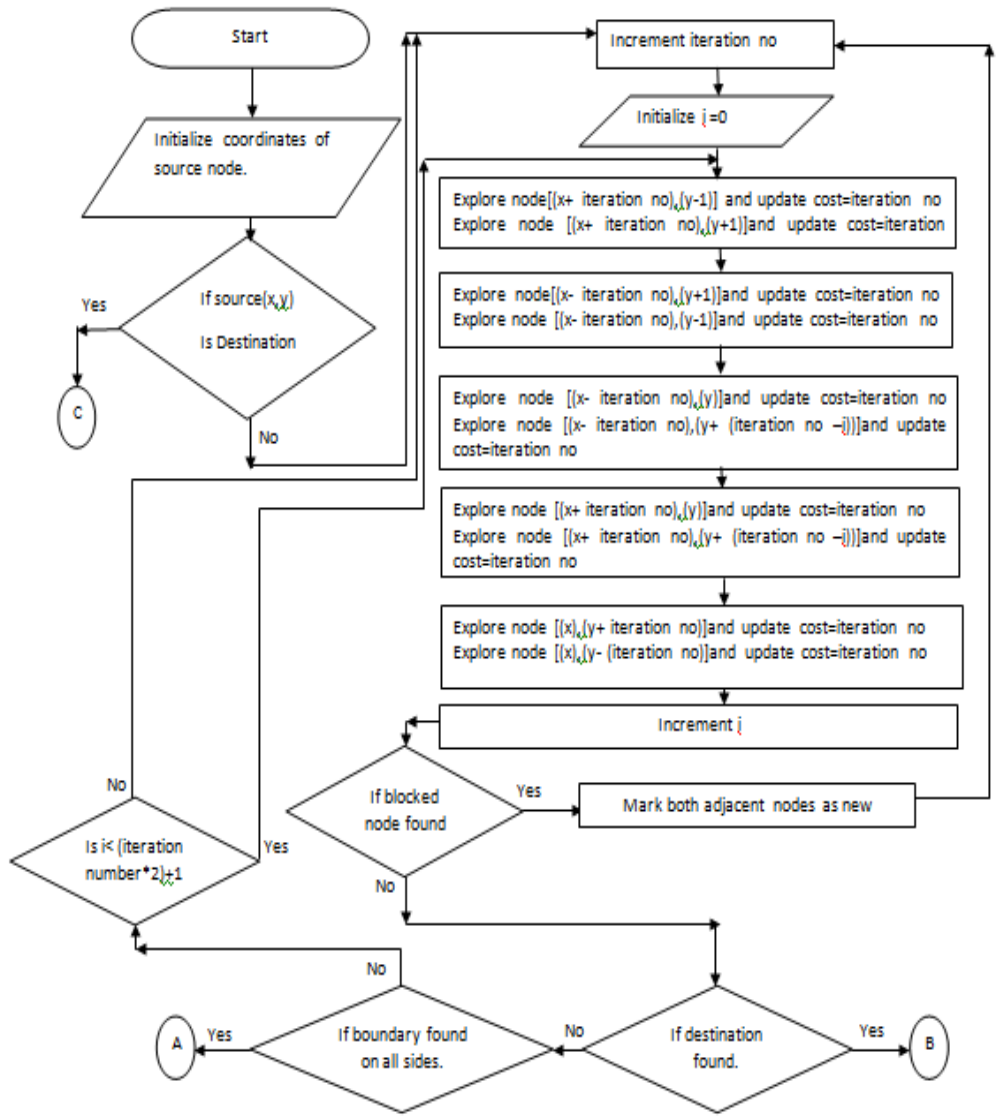

Figure 9. Flowchart for map exploring

Figure 10 shows flowchart for reverse traversing that is traversing back from destination node to source node in order to find shortest path. As seen in Figure 1 cost of each node is updated during traversal. This cost is used for backtracking i.e. finding shortest path. To find shortest path backtracking is done from destination to source by selecting nodes with minimum cost than others. If ever there are two nodes with same minimum cast then there exists more than one shortest path.

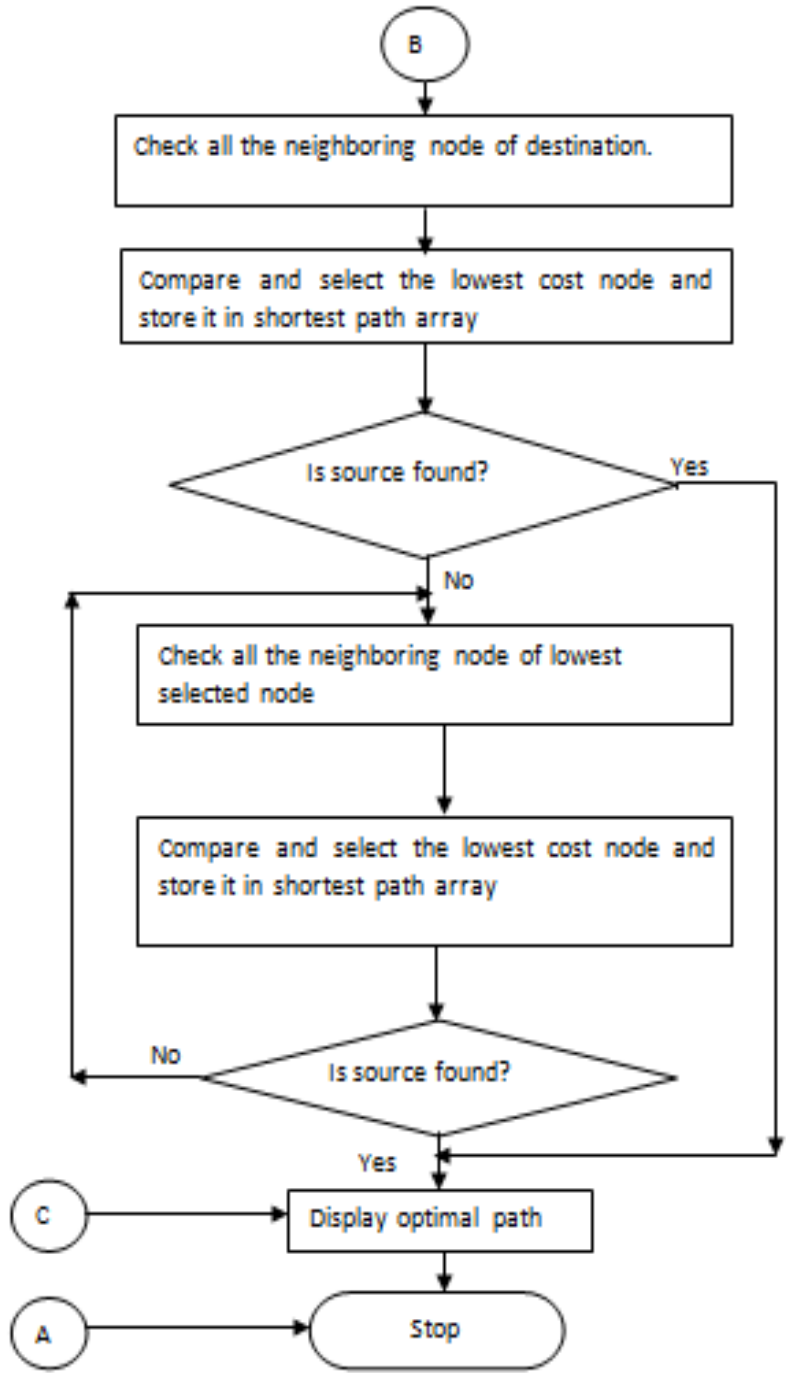

Figure 10. Flowchart for Reverse traversing (Shortest path)

# 4. RESULTS

Here results are analyzed and comparison of existing algorithm with proposed algorithm in terms of time complexity is shown.

## 4.1. Analysis of result

### 4.1.1. Map exploring in simple case (without obstacle)

The successive map exploring steps of A* and proposed algorithm in simple case where there no obstacles and few map exploring steps of A* and proposed algorithm in complex case in which obstacles are defined along the path.

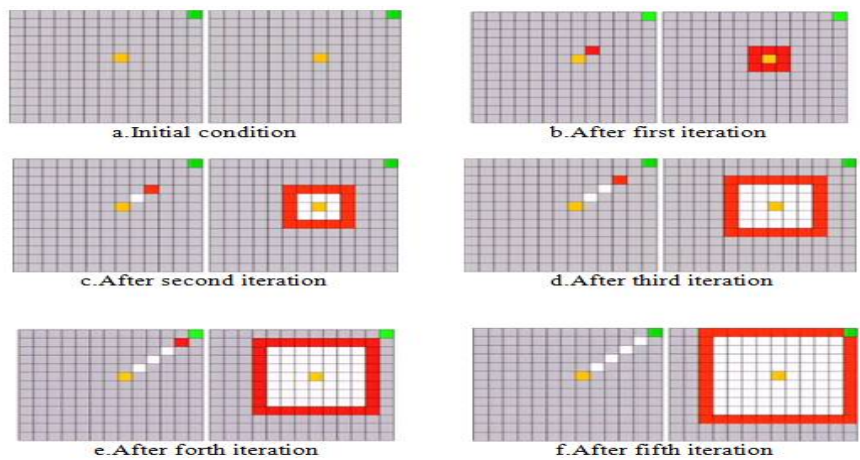

Figure 11. Map exploring (Simple case without obstacle)

Figure 11 shows successive map exploring steps of A* and proposed algorithm in simple case where there no obstacles. The first map (left-side) shows traversal by A* algorithm and adjacent figure (right-side) shows traversal by proposed algorithm. In above figure, yellow square indicates source node, green square indicates destination node, white squares indicates path traversed by respective algorithm, red squares indicate nodes traversed by current iteration. The proposed algorithm traverses in all four directions in one iteration whereas A* traverses as per Euclidean heuristic. In simple case number of iteration required by both algorithms is same.The detailed description of above figure is given below:

a: Initial condition is shown in which source and destination is defined.
b: In first iteration A* traverses node as calculated by Euclidean heuristic whereas in proposed algorithm 8 nearest neighbour of source is traversed and cost of all these nodes is updated as 1.
c: The traversal of A* is continued as per Euclidean heuristic and in proposed algorithm the second level neighbours from source is traversed.

Similarly the traversal of both the algorithm is shown in consecutive iteration. At fifth iteration both the algorithm reaches destination. In proposed algorithm, the reverse traversal is done to find shortest path here after.

**4.1.2: Map Exploring (With obstacles)**

Figure 12 shows few map exploring steps of A* and proposed algorithm in complex case in which obstacles are defined along the path.

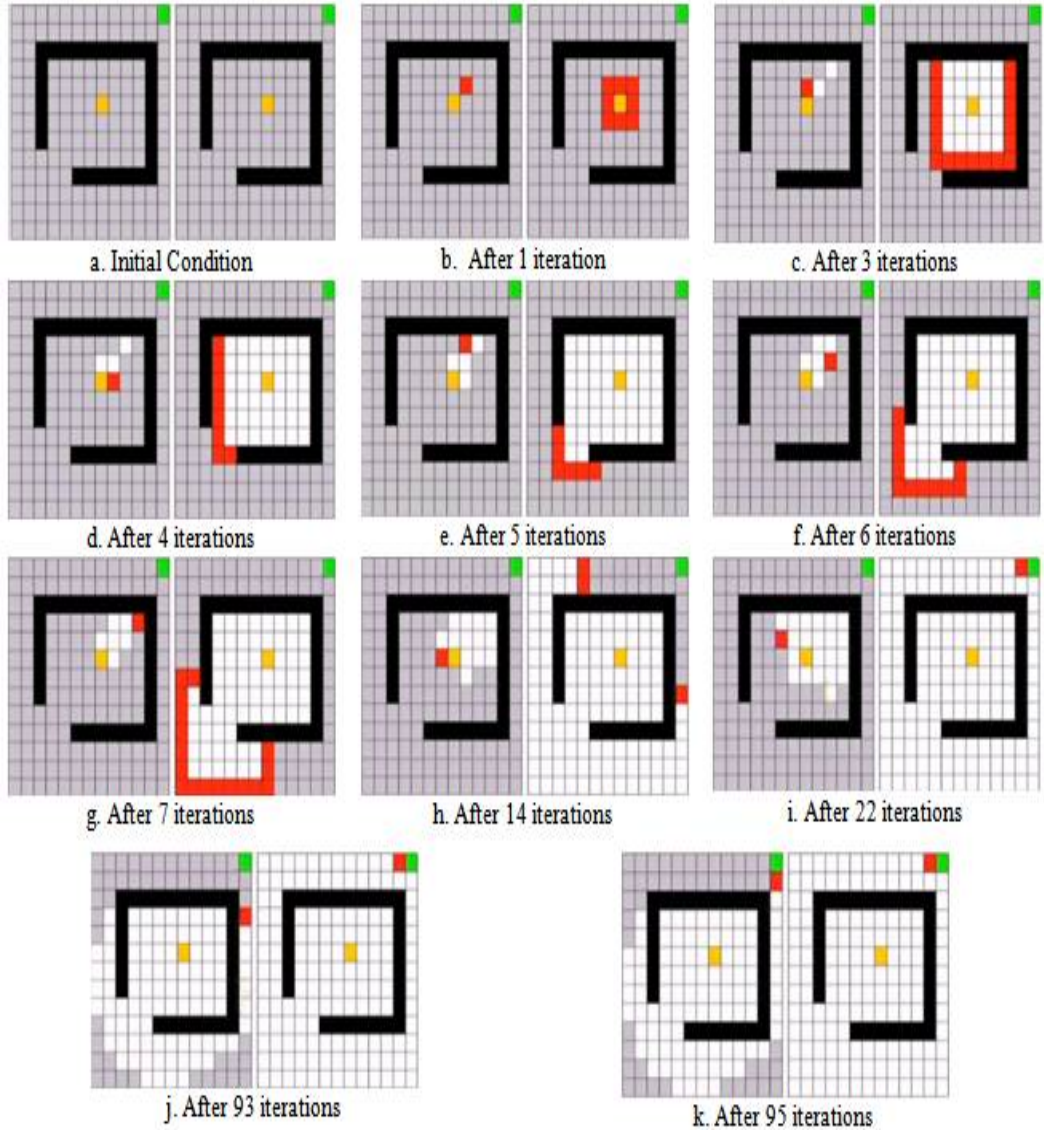

a. Initial Condition    b. After 1 iteration    c. After 3 iterations

d. After 4 iterations    e. After 5 iterations    f. After 6 iterations

g. After 7 iterations    h. After 14 iterations    i. After 22 iterations

j. After 93 iterations    k. After 95 iterations

Figure 12. Map exploring (With obstacles)

In above figure, black squares indicate obstacles. In this case, proposed algorithm reaches destination in 22 iterations (i.e. Figure 12.i ) whereas A* reaches destination in 95 iteration (i.e. Figure 12.k). Thus proposed algorithm reaches destination faster than A* algorithm. The detailed description of above figure is given below:

a: Initial condition in which obstacles, source and destination are defined in both cases.
b: Nodes traversed in first iteration by both the algorithm is shown. A* traverses as per Euclidean heuristic. The proposed algorithm traverse in all eight direction of source node.

c: The traversal by both the algorithm is continued in respective manner.

i: The proposed algorithm has reached destination. On the other hand A* has not yet reached destination and thus it continue the traversal.

j: The traversal of A* is shown. The proposed algorithm has already reached destination in 22 iterations.

f: A* reaches destination in 95 iterations.

## 4.2. Comparison with existing algorithm

Table 1 shows comparative study of proposed algorithm and existing algorithm. From the above table it is observed that the proposed algorithm has less time complexity than other algorithms.

Table 1. Comparitative Analysis

| Algorithm | Time Complexity | | Abbreviation |
|---|---|---|---|
| | With Obstacle | Without Obstacle | |
| Dijktra's Shortest Path | $O(|N|^2)$ | $O(|N|^2)$ | N – Number of nodes |
| A* (Based on heuristic) | $O(b^d)$ | $O((b^d) - (b*d))$ | b - Branching factor<br><br>d- Number of steps required to reach destination |
| Proposed Algorithm | $O(b*n + n)$ | $O(2n)$ | b – Number of obstacles in the path.<br><br>n – Number of steps required to reach destination. |

## 5. CONCLUSION

The proposed path finding algorithm reduces number of iteration required to traverse the path, thus reducing time complexity. Moreover the proposed algorithm comprises of a new method which populates the map faster followed by backtracking returning an optimized path without requiring destination co-ordinates in computing path.